\documentclass[10pt,twocolumn,letterpaper]{article}

\usepackage{cvpr}              
\usepackage[dvipsnames]{xcolor}

\usepackage{textcomp}
\usepackage{gensymb}
\usepackage{subcaption}
\usepackage{url}
\usepackage{xcolor}
\usepackage{colortbl}
\usepackage{makecell}

\makeatletter

\def\eg{\emph{e.g}\onedot} 

\def\ie{\emph{i.e}\onedot}

\makeatother

\newcommand{\myparagraph}[1]{\vspace{2.0pt}\noindent\textbf{#1.}}

\definecolor{newcolor}{rgb}{.8,.349,.1}
\definecolor{tablegrey}{rgb}{.61, .61, .61}

\usepackage{nicefrac}
\usepackage{multirow}
\usepackage{adjustbox}
\usepackage{arydshln}

\usepackage[accsupp]{axessibility}

\definecolor{cvprblue}{rgb}{0.21,0.49,0.74}
\usepackage[pagebackref,breaklinks,colorlinks,citecolor=cvprblue]{hyperref}
\usepackage{soul}
\DeclareRobustCommand{\hlgrey}[1]{{\sethlcolor{tablegrey!20}\hl{#1}}}

\def\paperID{3} 
\def\confName{FedVision}
\def\confYear{2024}

\title{Collaborative Visual Place Recognition through Federated Learning}

\author{Mattia Dutto\footnote{Corresponding author}, Gabriele Berton, Debora Caldarola, Eros Fanì, Gabriele Trivigno, Carlo Masone\\
Politecnico di Torino\\
{\tt\small name.surname@polito.it}\\
{\small $^*$Corresponding Author: Mattia Dutto}
}

\begin{document}
\maketitle
\begin{abstract}
Visual Place Recognition (VPR) aims to estimate the location of an image by treating it as a retrieval problem. VPR uses a database of geo-tagged images and leverages deep neural networks to extract a global representation, called descriptor, from each image. While the training data for VPR models often originates from diverse, geographically scattered sources (geo-tagged images), the training process itself is typically assumed to be centralized. This research revisits the task of VPR through the lens of Federated Learning (FL), addressing several key challenges associated with this adaptation. VPR data inherently lacks well-defined classes, and models are typically trained using contrastive learning, which necessitates a data mining step on a centralized database. Additionally, client devices in federated systems can be highly heterogeneous in terms of their processing capabilities. The proposed FedVPR framework not only presents a novel approach for VPR but also introduces a new, challenging, and realistic task for FL research, paving the way to other image retrieval tasks in FL.
\end{abstract}

\section{Introduction}
\begin{figure}[t!]
    \centering
    \includegraphics[width=\columnwidth]{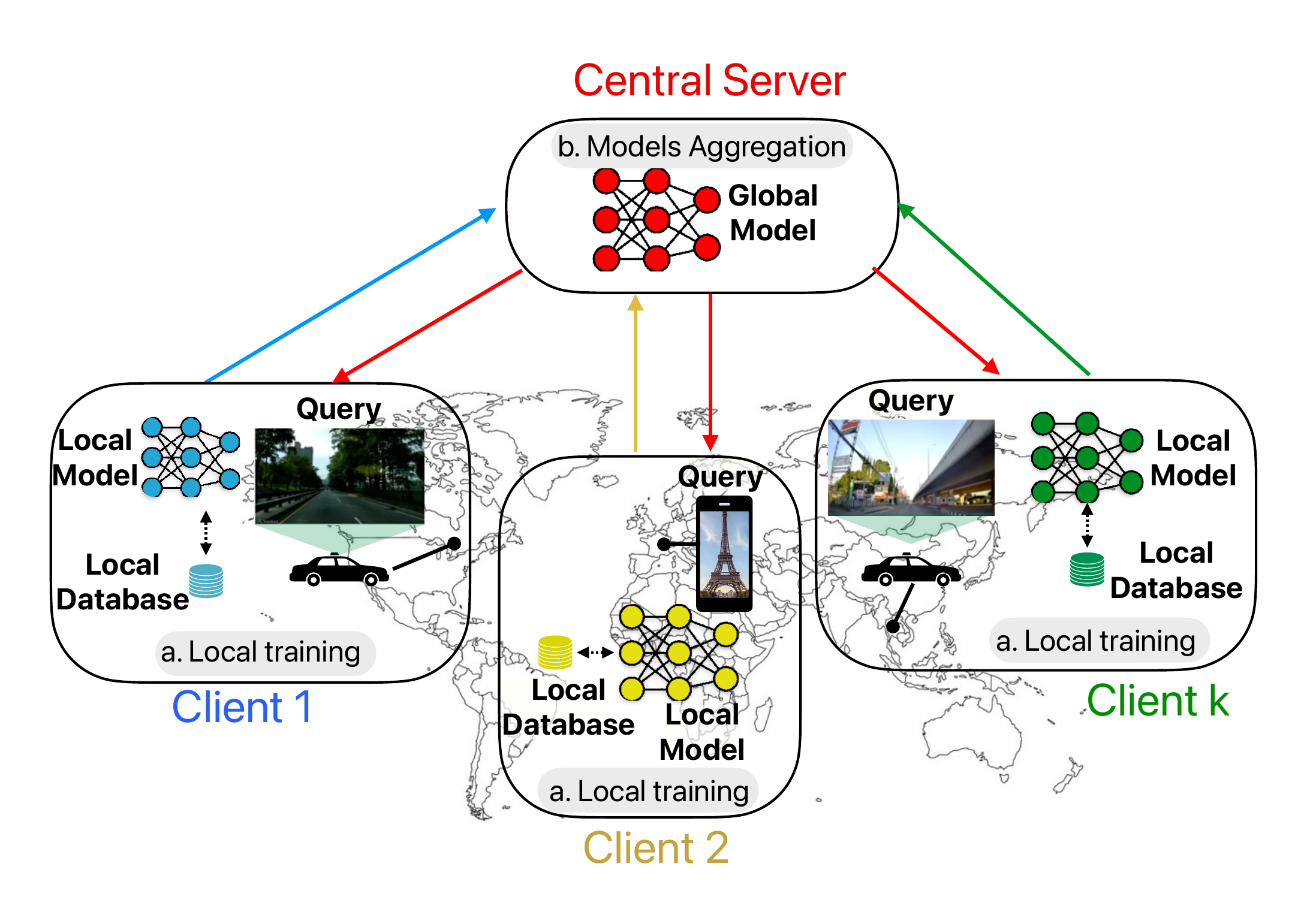}
    \caption{\textbf{Federated Visual Place Recognition} (FedVPR): we revisit the training of Visual Place Recognition models from the perspective of Federated Learning, with clients distributed across geographical areas, each possessing heterogeneous computational and communication resources and availability. Instead of relying on a central database for mining, each client builds its own database of geo-tagged images and uses it for local training based on contrastive learning (\hlgrey{step a.}). Subsequently, it communicates its model weights to the server, where they are aggregated into a new global model (\hlgrey{step b.}).} 
\label{fig:teaser}
\end{figure}

The ability to recognize the place depicted in a picture is of the utmost importance for many modern applications performed by camera-equipped mobile systems.
For example, in autonomous driving and mobile robotics, this ability is used for localization in instances where GPS measurement is unavailable or unreliable~\citep{Liu-2023_CMLocate, Suomela-2023_PlaceNav}, or in facilitating loop closure within SLAM (Simultaneous Localization and Mapping) pipelines~\citep{Lajoie-2023_SwarmSLAM}. Additionally, mobile phone applications heavily rely on this functionality for tasks like scene categorization~\citep{Apple-memories} and augmented reality support~\cite{Sarlin-2022_AR}. Likewise,  wearable devices leverage this capability to provide useful information to the user~\citep{Cheng-2020_wearable}. 
From a technical perspective, this task is referred to as \emph{Visual Place Recognition} (VPR)~\cite{Masone_2021_survey} and is naturally framed as an image retrieval problem. The query image to be localized is compared via features-space k-nearest neighbor (kNN) \cite{peterson2009k} to a database of images representing the known or already-visited places. Given that the database samples are usually labeled with geo-tags (such as GPS coordinates), the most similar images retrieved from the database serve as hypotheses of the queried location.
This approach entails representing each image with a single vector (\emph{global feature descriptor}) so that the kNN can efficiently compute the similarity between two images, \emph{e.g.}, as an Euclidean distance. 

Recent research on VPR has been focusing predominantly on the development of deep neural networks capable of extracting global feature descriptors that are both compact and highly informative for place recognition while leveraging large collections of data from highly heterogeneous distributions~\citep{Berton_2022_cosPlace, Alibey_2022_gsvcities, Berton_2023_EigenPlaces, Alibey_2023_mixvpr, Leyvavallina_2023_gcl,izquierdo_2023_salad}.
However, this centralized formulation assumes the images are readily available on one computer or a central server, which does not suit the distributed nature of the VPR applications previously discussed well. 
In an ideal scenario where mobile phones, wearable devices, and autonomous vehicles are deployed across numerous cities globally, it becomes crucial to leverage images collected by these diverse distributed devices without transferring their data to a central server, both for cost and privacy-related reasons. Furthermore, it would be beneficial to leverage the onboard computational capabilities of these devices to aid in model training. 

In light of these considerations, in this work we question \emph{how to revisit the training of VPR models from the perspective of Federated Learning} (FL)~\citep{mcmahan2017communication}, a distributed paradigm where multiple devices (\emph{i.e.}, clients) exchange model parameters updates with a central server to learn a shared global model, without any transfer of local data (see \cref{fig:teaser}).
Some notable challenges make the adaptation of VPR to FL not trivial. Unlike the conventional FL literature that revolves around classification problems \cite{caldas2018leaf,reddi2020adaptive,caldarola2022improving}, VPR lacks a clear division of data into classes. Instead, the collected images are labeled with continuous space annotations (commonly in the form of GPS coordinates), and models are usually trained with contrastive learning techniques~\citep{Arandjelovic_2018_netvlad}, which are often performed in conjunction with computationally heavy mining over a large centralized database \cite{Arandjelovic_2018_netvlad, Kim_2017_crn, Liu_2019_sare, Ge_2020_sfrs}:
in a federated setting, this would be unfeasible due to (i) the low computational capacity of the clients and (ii) the privacy concerns that a centralized database would create. 
By addressing these challenges, this paper introduces \textbf{FedVPR}, the first formulation of VPR in a federated learning paradigm. 

\textbf{Contributions}:
\begin{itemize}
    \item We introduce the first formulation of the VPR task in a federated learning framework. The importance of this formulation is twofold: for the VPR field, it opens up a new research direction with important practical implications; for the FL field, it provides a new downstream task that can broaden the horizon of the research community. 
    \item We propose a new splitting of the worldwide Mapillary Street-Level-Sequences (MSLS) dataset~\cite{Warburg_2020_msls} into federated clients, designed to replicate realistic scenarios with varying degrees of statistical heterogeneity across them.
    \item We deal with clients' data heterogeneity through critical design decisions such as client split, local iteration scheduling, and data augmentation, achieving centralized-level performances while accounting for power and computational requirements.
\end{itemize}

\section{Related work}

\myparagraph{Visual Place Recognition} (VPR) aims to geolocate a given input photo, called \emph{query}, by comparing it to a set of geo-tagged images (\emph{i.e.}, with known GPS position), called \emph{database}~\cite{Masone_2021_survey}.
Modern VPR methods leverage deep neural networks to extract global feature descriptors that provide a compact representation to perform a similarity search. An important milestone in this sense is the work by Arandjelović et al. ~\cite{Arandjelovic_2018_netvlad}, which introduced a learnable aggregation layer called NetVLAD and a training protocol that leveraged street-view imagery through a triplet loss.
The triplet loss paradigm and NetVLAD layer have been used with slight modifications in a number of successive papers \citep{Kim_2017_crn, Liu_2019_sare, Ge_2020_sfrs, Berton_2021_adageo, Mereu_2022_seqvlad, Hausler_2021_patch_netvlad}.
Since then, various innovations have been proposed, \emph{e.g.}, in the aggregation layers \cite{babenko2015aggregating,razavian2016visual, Alibey_2023_mixvpr, Radenovic_2019_gem, Tolias_2016_rmac, Zhu_2023_r2former}, architecture \cite{Zhu_2023_r2former,Kim_2017_crn,Wang_2022_TransVPR}, inference protocol~\cite{trivigno2023divide,arcanjo2022efficient}, training procedures \cite{Berton_2022_cosPlace, Leyvavallina_2023_gcl,Berton_2023_EigenPlaces,Alibey_2022_gsvcities,izquierdo2023optimal,Ge_2020_sfrs}, adaptation techniques \cite{Berton_2021_adageo, anoosheh2019night}, post-processing strategies \cite{Zhu_2023_r2former,Hausler_2021_patch_netvlad,Berton_2021_geowarp,Wang_2022_TransVPR}, use of foundational models \cite{izquierdo2023optimal, keetha2023anyloc}, as well as exploitation of temporal information \cite{zhao2024seq, garg2021seqnet, garg2021seqmatchnet, Mereu_2022_seqvlad, Berton2024_jist}.

Recently, the release of increasingly large datasets~\cite{Warburg_2020_msls,Berton_2022_cosPlace,Alibey_2022_gsvcities} and the recognition that the expensive mining and large outputs of the traditional "triplet-loss plus NetVLAD" paradigm hampers scalability~\cite{Berton_2022_benchmark} has led to the emergence of solutions that can learn more efficiently from the data. 
For example, CosPlace \citep{Berton_2022_cosPlace} and its derivative works \citep{Berton_2023_EigenPlaces,Berton2024_jist} use a mining-less classification proxy for training, whereas GCL~\citep{Leyvavallina_2023_gcl} overcomes the expensive mining by leveraging graded similarity labels and integrate this with a generalized contrastive loss. Conv-AP \citep{Alibey_2022_gsvcities} and MixVPR \citep{Alibey_2023_mixvpr} instead rely on a multi-similarity loss with more efficient online mining enabled by a curated dataset. 
Despite their strong performance, these methods are unsuitable for federated learning because they either (i) require the full database to initialize training \citep{Berton_2022_cosPlace,Berton_2023_EigenPlaces}, (ii) require batch sizes of hundreds of images from a curated dataset to converge \citep{Alibey_2022_gsvcities,Alibey_2023_mixvpr}, (iii) rely on large models and need additional similarity labels \citep{Leyvavallina_2023_gcl}, or (iv) need a re-ranking step \citep{Hausler_2021_patch_netvlad, Zhu_2023_r2former, Barbarani_2023_CVPR,Berton_2021_geowarp}, all too expensive to be performed on the clients.

These factors make it challenging to apply federated learning to VPR. Indeed, there are a few works that \emph{deploy} a single VPR model among multiple agents, which then localize collaboratively by fusing their descriptors~\cite{Li-2023_collaborative} or predictions~\cite{Lajoie-2023_SwarmSLAM} in a consensus-like mechanism, but without any local or coordinated training. To the best of our knowledge, this is the first work that uses a federated learning approach to \emph{learn} a global VPR model.

\myparagraph{Federated learning}\label{sec:rw_federated_learning} Federated Learning (FL) \citep{mcmahan2017communication}  enables learning from private data at the edge and is based on the exchange of model parameters between multiple clients and a central server over several communication rounds. Each client trains a local copy of a common global model independently on its own private data and only sends back the updated parameters, which are then aggregated on the server side. The server remains unaware of any sensitive information, safeguarding privacy. The versatility of FL has led to its successful application in various domains, ranging from medical imaging  \citep{feddg} to autonomous driving \citep{fantauzzo2022feddrive, fani2023feddrive, ladd} and natural language processing \citep{liu2021federated, lin2021fednlp, zhu2020empirical}.  
Despite these accomplishments, the exploration of FL's potential in more complex vision applications remains incomplete \citep{caldarola2022improving}. 
This study marks a significant step in advancing FL, presenting the first reexamination of VPR within the federated setting. Adapting VPR to FL poses a major challenge due to the task's reliance on mining large datasets. Furthermore, as VPR does not hinge on a discrete label space — unlike the majority of FL literature centered around classification methods \cite{caldas2018leaf,caldarola2022improving} — this work introduces a novel case study, broadening the horizons of the FL research community. Aiming at studying real-world scenarios, we also take into consideration the challenges arising in realistic federated settings, namely \textit{statistical} and \textit{system heterogeneity} \citep{Li_2020,zhu2021federated,reddi2020adaptive}. Differently from standard distributed training scenarios, the clients are heterogeneous in terms of the distribution of the local data and computational capabilities and availability over time. This heterogeneity may arise from factors such as the users' geographical locations or differences in internet access. 
Due to such distribution shifts, the learning trend becomes inherently noisy and unstable \citep{acar2021federated,reddi2020adaptive,caldarola2022improving} and achieving a target performance necessitates more training rounds, impacting the communication network. Several approaches tackle this issue. Some focus on regularizing local training to prevent local objectives from deviating significantly from the global one  \citep{li2020federated,karimireddy2021scaffold,acar2021federated}. Others aim to virtually equalize the number of samples across clients (\eg, FedVC \cite{hsu2020federated}) or group similar clients together \citep{sattler2020clustered,ghosh2020efficient,duan2021flexible,caldarola2021cluster}. 
Following the latter line of works, we further investigate hierarchical FL \citep{briggs2020federated,chu2023design} for VPR. Clients found in close geographical locations are grouped together, with each cluster referring to a distinct server. All the servers communicate with a first-level additional server, orchestrating the overall training process. This approach facilitates a scalable VPR system that can adapt to regional preferences while maintaining a globally consistent framework. 
\section{Method}

\begin{figure*}
    \centering
    \includegraphics[width=.9\linewidth]{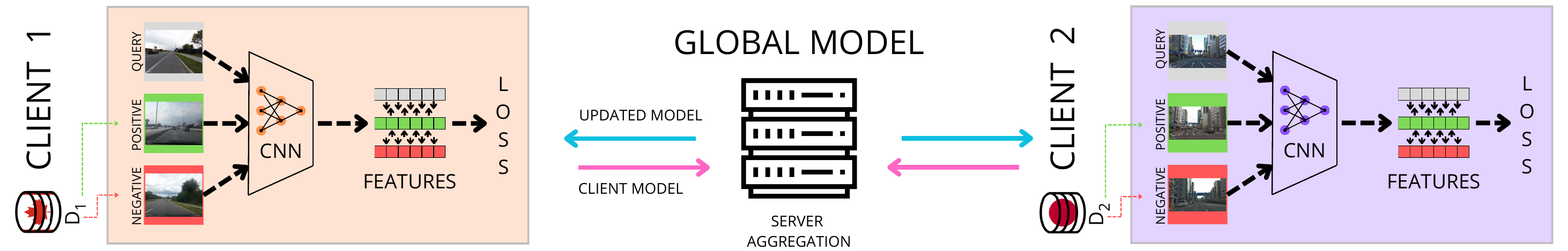}
    \caption{\textbf{FedVPR training.} At each round $t$, the server \textcolor{Turquoise}{sends the current global model} to a set of active clients, \eg \textit{client 1} and \textit{client 2} in the figure. Each client $i$ has access to its own local dataset $\mathcal{D}_i$, whose distribution is highly influenced by the user's geographical positions (hence the country flags on the local datasets). Differently from centralized VPR, in FedVPR the mining happens exploiting the client's previously collected images. Thus, given a \textcolor{Gray}{query image}, local optimization is based on a contrastive loss, which relies on a \textcolor{ForestGreen}{positive} and \textcolor{BrickRed}{negative} images extracted from $\mathcal{D}_i$. Since each local dataset follows a different distribution, the resulting updated parameters vary from client to client (\textcolor{Peach}{\textit{orange}} vs. \textcolor{Orchid}{\textit{purple}} updates). Lastly, the \textcolor{Rhodamine}{local parameters are sent back to the server}, where they are aggregated with FedAvg.}
\label{fig:network}
\end{figure*}

\subsection{Centralized VPR}
\label{sec:centr_vpr}

The task of VPR is commonly approached as an image retrieval problem. Given a query image, the goal is to find the most similar matches within a geo-tagged database to infer the query's location.
When training a VPR model, we aim to learn a function $F_\theta : \mathbb{X} \rightarrow \mathbb{D}$ parameterized by $\theta\in\mathbb{R}^p$ which projects each image sample $x \in \mathbb{X}$ into a common embedding space $\mathbb{D}$ of dimensionality $d$.
Intuitively, $F_\theta$ should provide an embedding space where different representations of the same place (\emph{e.g.}, the same building seen with different perspectives or illuminations) should be close to each other while simultaneously being far from representations of other places.
VPR models are commonly trained with contrastive losses \cite{Arandjelovic_2018_netvlad,Ge_2020_sfrs, Zaffar_2021_vprbench, Berton_2022_benchmark}, which rely on feeding the model with 
samples from the same place (\textit{positives}) and samples from different, although potentially similar, locations (\textit{negative}).
The most common approach is to use a triplet loss \cite{Arandjelovic_2018_netvlad}, which takes a query (\textit{anchor}), a positive and a negative image, and aims at bringing the query and the positive sample closer in the features space, while pushing away the negative one.
However, when using such formulations of the loss, if the chosen negative's embeddings are already far away from the query's ones (a \textit{trivial negative}), the loss will be close to zero, thus leading to uninformative gradients. 
To circumvent this issue, hard negatives (\emph{i.e.}, negatives close to the query in features space, or visually similar)
must be selected, enabling the model to reach higher performances \citep{Berton_2022_benchmark}.
The hard negatives selection process takes the name of \textbf{mining} and is a time-consuming technique performed throughout the training phase to select ever-increasing difficult negatives for each given query.
Formally, for a given training query $q\in\mathbb{X}$, we want to obtain a training triplet $(q, \tilde{p}^q, \tilde{n}^q)$, where $\tilde{p}^q$ is its positive image and $\tilde{n}^q$ the negative one. 
The set of potential positives $ \mathcal{P}:= \{p^q_i\} $, commonly defined as the images within a threshold $\tau=25$ meters from the query \cite{Arandjelovic_2018_netvlad, Kim_2017_crn, Zhu_2023_r2former}, is retrieved using the GPS label.
The set of negatives $\mathcal{N} := \{n^q_i\}$ instead contains all the images geographically far from the query and is obtained following the symmetrically opposite approach.
Since GPS labels alone are not enough to determine whether an image actually depicts the same visual content (\emph{e.g.}, close-by images could point in opposite directions), the current estimate of $F_\theta$ is used to compute the Euclidean distance $D_\theta(q, p^q_i)$.
Thus, the candidate with the highest probability of being a true positive is selected according to the criteria
\begin{equation}
    \tilde{p}^q = \underset{p^q_i\in\,\mathcal{P}}{\text{argmin}}~ D_\theta\left(q, p^q_i\right).
\end{equation}
On the other hand, hard negatives are selected as those who are closest to the query in the embedding space (therefore visually similar), while still being geographically far.
\begin{equation}
    \tilde{n}^q = \underset{n^q_i\in\,\mathcal{N}}{\text{argmin}}~ D_\theta\left(q, n^q_i\right).
\end{equation}
Finally, the loss for the selected triplet $(q, \tilde{p}^q, \tilde{n}^q)$ is
\begin{equation}
    \mathcal{L}_\theta = \max\left( D^2_\theta\left(q, \tilde{p}^q\right) - D^2_\theta\left(q, \tilde{n}^q\right) + m, 0\right),
\end{equation}
where $m$ is the margin hyperparameter. \\

Finally, while the above procedure (triplet loss with negative mining) is not the only way to train VPR models, its computational affordability makes it suitable for a federated learning scenario. On the contrary, latest methods that skip mining altogether either require to know the entire database surface area before starting training \citep{Berton_2022_cosPlace}, or are computationally expensive due to large batch sizes \citep{Alibey_2022_gsvcities, Alibey_2023_mixvpr} or large models \citep{Leyvavallina_2023_gcl}, both of which are unsuitable options within a federated scenario.


\subsection{Federated Visual Place Recognition (FedVPR)}
\label{sec:fedvpr}
In this work, we consider a realistic scenario in which the goal is to build a VPR system that will be deployed to a set of clients. While it is expected that some data will be available to pre-train a model in a centralized fashion, in general, it is not easy to obtain large-scale annotated datasets for place recognition. As an example, while large datasets exist available for research purposes, they are usually scraped from Google Street View, which does not allow commercial usage, and collecting a dataset that covers large geographical areas independently is highly expensive. On the other hand, the clients to which the model is deployed are a convenient source of heterogeneous and relevant data, which are fundamental for training a robust feature extractor for the task at hand. However, in the spirit of Federated Learning, the client's privacy must not be violated. Possible examples of the described scenario can be a company deploying a fleet of self-driving vehicles, wanting to improve the performances of its localization-and-mapping pipeline, a swarm of drones, or even general-purpose content-based image retrieval. We frame our analysis in this realistic scenario, demonstrating the criticalities of developing a distributed learning framework. \cref{fig:network} summarizes FedVPR.

\myparagraph{Federated framework} In the standard federated framework, a central server communicates with a set of clients $\mathcal{C}$ over $T$ communication rounds. Clients commonly are edge devices, \emph{e.g.}, smartphones, autonomous vehicles, and IoT sensors. In our setting, each client $k \in \mathcal{K}$ has access to a privacy-protected dataset $\mathcal{D}_k$ made of $N_k$ images $x\in\mathbb{X}$ associated with a GPS location. The training goal is to learn a global shared model $F_\theta:\mathbb{X} \rightarrow \mathbb{D}$ parameterized by $\theta\in\mathbb{R}^p$ without violating the users' privacy. At each round $t$, the server sends the current global model $\theta^t$ to a subset of selected clients $\mathcal{C}^t\in\mathcal{C}$, which trains it using their local data. 
\myparagraph{Local mining} As detailed in \cref{sec:centr_vpr}, the \textbf{mining} process is crucial for learning as it ensures that the network is fed with informative samples during training. 
Differently from the centralized scenario where the model has access to the whole dataset for training, here we encounter the challenge of performing mining without (\textit{i}) increasing the communication costs by exchanging continuous information between clients and server, \textit{ii}) downloading enormous quantities of data on resource-constrained devices and (\textit{iii}) exchanging data with other clients or the server, which could result in privacy leaks. In FedVPR, to avoid the aforementioned bottlenecks and any privacy concerns, the mining is limited to the database images previously collected by each client. The local data collection likely satisfies the requirement of having access to \textit{hard negative} samples (\emph{i.e.}, visually similar images of different places) to successfully train the feature extractor. However, its limited variability is an important factor that can slow down convergence and ultimately affects performances. More details on this matter are presented in the experimental section. 

At the end of local training, each client $k$ sends the updated $\theta_k^t$ to the server. 
The global training objective is solved by aggregating the received updates on the server side. The de-facto standard aggregation algorithm is FedAvg \cite{mcmahan2017communication}, which averages the clients' parameters as 
\begin{equation}
    \theta^{t+1} \leftarrow \sum_{k\in\mathcal{C}^t} \frac{N_k}{N} \theta^t_k,
    \label{eq:fedavg}
\end{equation}
where $N=\sum_{k\in\mathcal{C}^t} N_k$. As showed by \cite{reddi2020adaptive}, \cref{eq:fedavg} is equivalent to apply SGD (Stochastic Gradient Descent) \cite{ruder2016overview} with a pseudo-gradient $\Delta \theta^t := \sum_{k\in\mathcal{C}^t} \nicefrac{N_k}{N} (\theta^t - \theta_k^t)$ and server learning rate $\eta_s=1$, where each round is a different optimization step. Thus, \cref{eq:fedavg} can be generalized to $\theta^{t+1} \leftarrow \theta^t - \textsc{ServerOpt}(\theta^t, \Delta \theta^t, \eta_s, t) $, with \textsc{ServerOpt} being any gradient-based optimizer.

The realistic scenario considered in this work gives us the opportunity to study not only the challenges linked with the VPR task but also those derived from the federated framework. In particular, in the experiments discussed in  \cref{sec:exp}, we investigate the effect of \textit{statistical heterogeneity} on the task of VPR in real-world settings where clients' data is usually non-\textit{i.i.d.} with respect to the overall data distribution. Namely, given two clients $i$ and $j$, their local datasets likely follow a different distribution $P$, \emph{i.e.}, $P_i \neq P_j$. Consequently, local optimization paths usually lead towards distinct local minima, straying from the global one, a phenomenon referred to as \textit{client drift} \cite{karimireddy2021scaffold}. The resulting learning trends are slowed down, unstable, and subject to catastrophic forgetting. 

\myparagraph{\textbf{Hierarchical FL}} Lastly, since in our setting, the clients' data distributions are linked with the users' geographical locations (\emph{e.g.}, a device likely spends most time within a single region), we additionally explore the hierarchical FL (H-FL) setup \cite{briggs2020federated}.  In H-FL, we group clients into $K$ clusters according to their geographical proximity. A specialized model $F_{\theta_c}$ is assigned to each cluster $c$. Once every $T_c$ rounds, the cluster-specific models are aggregated. This implies the existence of multiple servers. We explore a dual-level framework: the first-tier servers handle inter-clusters interactions (\emph{e.g.}, among cities or continents), while second-tier ones manage the intra-cluster exchanges (\emph{e.g.}, between users living in the same city, or continent). 

\section{Decentralizing the MSLS dataset for FL}
\label{sec:dataset}
Our experiments center on the Mapillary Street-Level-Sequences (MSLS) dataset \cite{Warburg_2020_msls}, geographically distributed across 30 cities worldwide, mimicking a FL scenario.
The dataset is split into non-overlapping train, validation, and test sets. Each set comprises distinct cities: Amsterdam and Manila for validation, San Francisco and Copenhagen for testing, and the remaining cities for training.
Similar to other VPR datasets, each subset is further divided into databases and queries. Queries represent images to be localized, while databases act as the system's prior knowledge of the area. Notably, the dataset excels due to its rich diversity. It encompasses a vast number of cities captured by various users, resulting in a wide range of cameras, weather conditions, times of day, and scenarios across both urban and rural environments. These characteristics perfectly align with the demands of our use case. 


\subsection{Proposed FL datasets}
\begin{table}
\caption{Characteristics of the proposed FL+VPR datasets and associated FedAvg performances.}
    \centering
    \begin{adjustbox}{width=\linewidth}
        \begin{tabular}{lccccc}
            \toprule
             \textbf{FL dataset} & \makecell{\textbf{Radius} \\ \textbf{(m)}} & \makecell{\textbf{Sequences} \\ \textbf{per client}} &  \makecell{\textbf{Images} \\ \textbf{per client}} & \makecell{\textbf{Number of} \\ \textbf{clients}} & \textbf{R@1 (\%)}  \\
            \midrule
            \rowcolor{tablegrey!20} Centralized & - & - & - & - & 66.0 \scriptsize{$\pm$ 0.4} \\ 
            \midrule
            Random & - & 64 \scriptsize{$\pm$ 1}& 3655 \scriptsize{$\pm$ 676}  & 700 & 40.2 \scriptsize{$\pm$ 0.0} \\
            \midrule
            Clustering & - & 36 \scriptsize{$\pm$ 32} & 2018 \scriptsize{$\pm$ 1266}& 678 & 57.3 \scriptsize{$\pm$ 1.2} \\
            \midrule
            \multirow{3}{*}{Proximity} & 1000 & 17 \scriptsize{$\pm$ 18} & 897 \scriptsize{$\pm$ 808} & 1303 & 51.7 \scriptsize{$\pm$ 1.7} \\
             & 2000 & 33 \scriptsize{$\pm$ 48} & 1834 \scriptsize{$\pm$ 2050}& 713 & 61.0 \scriptsize{$\pm$ 0.6} \\
             & 4000 & 75 \scriptsize{$\pm$ 148} & 4270 \scriptsize{$\pm$ 6515} & 316 & \textbf{66.1 \scriptsize{$\pm$ 0.3}} \\
            \bottomrule
        
        \end{tabular}
        
    \end{adjustbox}
    \label{tab:splits}
\end{table}
This work's first contribution lies in proposing three novel splits for the MSLS dataset. These splits mimic real-world scenarios with varying data distributions across devices. Users are grouped based on geographical \textit{proximity}, \textit{similarity} in city features (\eg, architecture), or \textit{randomness}. \cref{tab:splits} summarizes the datasets' characteristics. Additional analyses can be found in \cref{app:plots}.

\myparagraph{Proximity} This split emulates user movements within a neighborhood or a proximal geographical area. While clients in smaller towns may explore different localities, users in large cities like Tokyo or San Francisco are inclined to stay within their neighborhoods. 
The MSLS dataset is first divided geographically, with each city representing a separate entity. Within each city, clients are formed iteratively. An initial query image is chosen from a sequence available in that city. All other geographically close sequences, \ie, within a given radius from the coordinates of the selected image, are then grouped with the chosen query image. This group is considered a valid client only if it contains at least two queries and two database sequences. 
The resulting number of training clients depends on the chosen radius,  which we select in $\{1000, 2000, 4000\}$ meters. Twelve clients are randomly selected from the pool of validated training clients to serve as the validation set. The test set is kept on the server side.

\myparagraph{Clustering} The proximity split assumes similar features (\eg, architecture) in nearby areas. However, distant neighborhoods might share more similarities (\eg, busy streets, shops) than geographically close ones. To capture such nuances, the clustering split utilizes the $K$-means algorithm \cite{lloyd1982least} at the city level, grouping images based on their visual and environmental characteristics. To ensure a balanced number of clients while capturing similarities, we determine the value of $K$ (number of clusters) for each city individually. We use the number of clients obtained in the proximity split with a radius of 2000 meters as a reference point, \ie, $K=713$. The same selection criterion of the proximity split is then applied to define the valid clients. 12 clients are maintained for validation, and the test set is on the server.

\myparagraph{Random} Following the approach of \cite{hsu2019measuring,fantauzzo2022feddrive,fani2023feddrive}, we introduce a \textit{random} split of MSLS to emulate a uniform distribution and facilitate the understanding of the effects of statistical heterogeneity induced by domain shift. Each dataset's client includes images from all cities, and validation is conducted on the same local dataset. If a city does not have enough data for all clients, we duplicate existing sequences until each client can access at least one sequence from each city. Any remaining sequences are redistributed among clients. The test set remains on the server side. 

\section{Experiments and Results}
\label{sec:exp}

\begin{table}
    \caption{\textbf{Centralized baselines}. 
    Comparison of different model architectures (\textit{left}) and pooling layers (\textit{right}) in terms of recall. The average pooling layer is used for the architecture comparison on the left, and ResNet18 truncated on the right. }
    \centering
    \begin{subtable}{0.63\linewidth}
        \centering
        \begin{adjustbox}{width=\linewidth}
            \begin{tabular}{lccc}
                \toprule
                \textbf{Backbone} & \textbf{R@1} & \textbf{Total} & \textbf{Trained} \\
                \midrule
                ResNet18 trunc. & $42.9 \pm 2.5$ & \textbf{2.8M} & \textbf{2.1M} \\
                ResNet18 & $\textbf{60.1} \pm \textbf{0.3}$ & 11.2M & 10M \\
                VGG16  & $46.3 \pm 0.5$ & 14.7M & 7M \\
                \bottomrule
            \end{tabular}
        \end{adjustbox}
    \end{subtable}
    \hspace{.02\linewidth}
    \begin{subtable}{0.31\linewidth}
        \centering
        \begin{adjustbox}{width=\linewidth}
            \begin{tabular}{lc}
                \toprule
                \textbf{Pooling} & \textbf{R@1} \\ 
                \midrule
                SPOC \cite{babenko2015aggregating} & $42.9 \pm 2.5$ \\
                MAC \cite{razavian2016visual} & $59.4 \pm 0.6$ \\ 
                GeM  \cite{Radenovic_2019_gem} & $\textbf{68.0} \pm \textbf{0.3}$ \\
                \bottomrule
            \end{tabular}
        \end{adjustbox}
    \end{subtable}
    \label{tab:centralized-baseline}
\end{table}
\subsection{Implementation details}
\label{sec:impl_details}
This section provides the main implementation details used in our experiments. Additional information can be found in \cref{app:impl_details}. 
The used model architecture, unless specified otherwise, is a ResNet18  \cite{he2015deep} truncated at the third convolutional layer, with GeM pooling \cite{Radenovic_2019_gem}. 
In local training, we use a batch size of 2 triplets per iteration, with each triplet comprising 5 negative examples for each query alongside its positive counterpart. The learning rate is $1 \times 10^{-5}$ with Adam as both local optimizer and optimizer in the centralized runs. On the server side, unless otherwise specified, FedAvg \cite{mcmahan2017communication} is used for model aggregation. 
In the centralized experiments, the training continues until the model converges, incorporating an early-stopping mechanism based on validation accuracy. In contrast, in the FL framework, each round engages 5 clients, with each client running a single local epoch. This process is repeated across a total of $T=300$ rounds. Validation is conducted using a subset of 12 clients that do not participate in the training phase. Lastly, testing is directly handled by the server.

The Hierarchical Federated Learning (H-FL)  \citep{hfl, chu2023design} experiments propose two hierarchy types, delineated by geographical proximity: \textit{City} and \textit{Continental} levels, where clients within the same city or continent respectively are aggregated to form the cluster-specific models. This results in 21 clusters in the former case and 4 in the latter. We employ the classical SGD server optimizer in our hierarchical experiments and select 5 clients from each cluster per round. 

All experiments are conducted using an image resolution of $288 \times 384$ pixels, which provides a good trade-off between speed and results. Notably, we refrain from employing any form of data augmentation in our methodology. 
To ensure the robustness and reliability of the results, all experimental outcomes are averaged over 3 distinct runs. 


\begin{figure}[t]
    \centering
    \includegraphics[width=.75\linewidth]{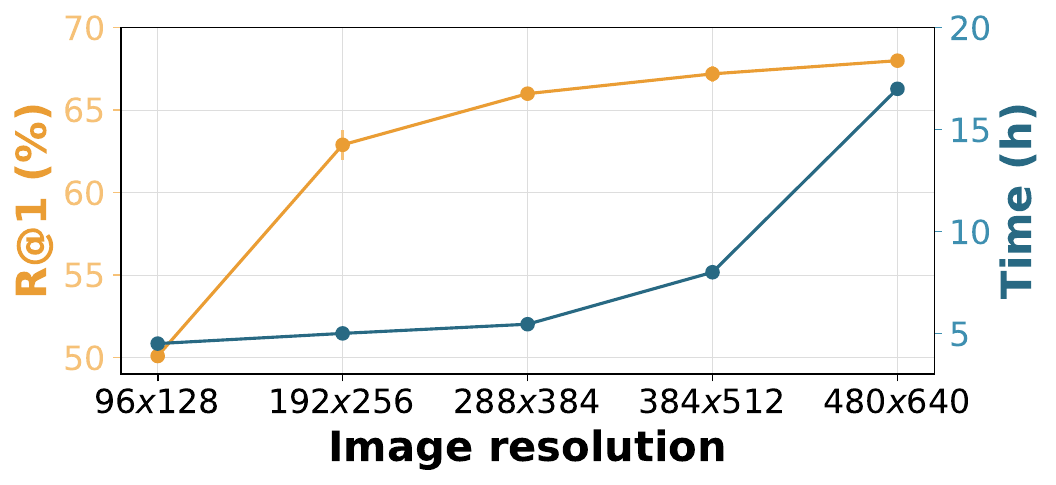}
    \vspace{-0.2cm}
    \caption{\textbf{Centralized setting}. Comparison of R$@1$ (\%) and computational time (hours) when varying the image resolution. Resolution greatly affects training time, and an optimal trade-off can be attained with minimal performance drops.
    }
    \label{fig:centralization-resolution}
    \vspace{-10pt}
\end{figure}

\subsection{Centralized baselines}
\label{exp:centr}
In FL, the choice of network architectures is restricted by the communication bottleneck and the constrained computational capabilities of individual clients. Consequently, lightweight backbones with fewer parameters are preferred to alleviate communication and computation burdens.
\Cref{tab:centralized-baseline} presents a comparison of backbone architectures and aggregation layers in a centralized setting, which serve as a baseline for the FL experiments. We consider three networks pre-trained on ImageNet~\cite{imagenet}: ResNet18 \citep{he2015deep} truncated at the third layer, ResNet18 with fine-tuning limited to the last block, and VGG16 \citep{simonyan2015deep}.
Regarding aggregation layers, we test different pooling strategies, namely SPOC \citep{babenko2015aggregating}, MAC \citep{razavian2016visual}, and GeM \citep{Radenovic_2019_gem}. These methods apply a pooling operation on the feature maps provided by the backbone, obtaining a single embedding for each image, whose dimensions are determined by the number of channels of the chosen backbone. NetVLAD \citep{Arandjelovic_2018_netvlad} is another popular aggregation layer for VPR, which usually grants robust performances \cite{Berton_2022_benchmark}. Nevertheless, some drawbacks make its adoption impractical in a federated scenario. Firstly, it outputs large descriptors (ranging from 16k to 64k dimensionality, depending on the backbone), which clients would have to store and use for nearest neighbor search. Moreover, it requires to initialize a set of centroids (\textit{visual words}) using a diverse set of images \cite{Berton_2022_benchmark}, which would have to be provided by the server, increasing the communication costs and eventually raising privacy concerns. Thus, in our analysis, we consider simpler pooling layers to be viable alternatives.

Based on the results in \cref{tab:centralized-baseline}, we select ResNet18 truncated as our architecture, capitalizing on the benefits of a reduced parameter count. The adoption of GeM as the pooling layer is justified by the observed performance improvements. However, GeM increases the computational time, which is a restriction in FL settings, where edge devices have access to limited resources.  To mitigate this, we compare various image resolutions, considering both training time and performance metrics in \cref{fig:centralization-resolution}. We settle on the 288$\times$384 resolution as the optimal choice, striking the best balance between recall and training time. With this resolution, we get a final centralized recall of $66.0 \pm 0.4$.


\subsection{FL baselines}
\label{sec:exp_fl}
\myparagraph{Splits comparison} With the centralized baselines established, we conduct an analysis of the vanilla FedAvg algorithm across the different introduced datasets, as illustrated in \cref{tab:splits}. 
Interestingly, the performance of FedAvg on the Random FL dataset significantly lags behind that of other ones, despite the scenario closely resembling uniform splits as seen in prior works \citep{hsu2019measuring, fantauzzo2022feddrive, fani2023feddrive}. However, as highlighted in \citep{Berton_2022_benchmark}, optimal performance necessitates hard negatives to be situated within a distance range of 25 meters to a few kilometers from the query, a condition not met in this dataset where images within the same client can belong to various locations worldwide. 
Our experiments on the Proximity and Clustering FL datasets yielded comparable performance. Interestingly, the Proximity split achieved slightly better results on average. This difference is likely because images within clients of the Proximity split have closer GPS coordinates compared to those in the Clustering split. Additionally, clients in the Proximity FL datasets with radii of 2000m and 4000m also contain a larger number of images compared to their counterparts in the Clustering datasets. 
The proximity experiment employing a radius of 4000m demonstrates performance levels roughly akin to the centralized baseline. However, opting for a larger radius results in fewer clients, each possessing a greater number of images and sequences, thereby resembling a cross-silo scenario \cite{li2020review}. Conversely, reducing the radius yields a larger number of clients but with a markedly limited quantity of images and sequences per client.
Given all these considerations, we shift our focus solely to the \textbf{proximity} FL dataset with a \textbf{2000m radius} in the upcoming experiments. This choice strikes a balance between the number of clients and the volume of data per client.

\begin{table}
    \caption{Comparison of the vanilla baseline FedAvg with Hierarchical FL methods and various server optimizers. Notation: \textbf{CC} for continent-level middle servers in H-FL, \textbf{C} for city-level middle servers, SGDm for SGD with server-side momentum, $T$ rounds, $C$ clients participating at each round.}
    \vspace{-0.4cm}
    \begin{center}
    \begin{adjustbox}{width=.8\linewidth}
    {{
        \begin{tabular}{lcc}
            \toprule
                \textbf{Algorithm} &\textbf{Server Optimizer} & \textbf{R@1 (\%)} \\
            \midrule
            \multirow{4}{*}{FedAvg} & SGD & 61.0 \scriptsize{$\pm$ 0.6} \\
            & SGDm & 61.2 \scriptsize{$\pm$ 1.4} \\
            & Adam & 61.1 \scriptsize{$\pm$ 1.2} \\
            & AdaGrad & 61.6 \scriptsize{$\pm$ 0.3} \\
            \midrule
            \multirow{1}{*}{H-FL (\textbf{CC})} & SGD & 46.9 \scriptsize{$\pm$ 1.3} \\
            \hdashline
            FedAvg $T=75$, $C=20$ & SGD & 55.6 \scriptsize{$\pm$ 0.8} \\
            \midrule
            \multirow{1}{*}{H-FL (\textbf{C})} & SGD & 33.3 \scriptsize{$\pm$ 0.1} \\
            \hdashline
            FedAvg $T=15$, $C=105$ & SGD & 44.2 \scriptsize{$\pm$ 0.4} \\
            \bottomrule
        \end{tabular}
    }}
    \end{adjustbox}
    \end{center}
    \label{tab:fl-cmp}
    \vspace{-20pt}
\end{table}

\myparagraph{Baselines} \Cref{tab:fl-cmp} presents a comparative analysis between FedAvg with various server-side optimizers and baselines sourced from the Hierarchical Federated Learning (H-FL) literature \citep{hfl, chu2023design}. As described in \cref{sec:impl_details}, we distinguish between H-FL at city (C) and continent level (CC). To ensure a fair comparison with H-FL, which selects 5 clients from each cluster per round, we additionally run FedAvg with 20 clients per round and $T=75$, and 105 participating clients for $T=15$ rounds. 
Both H-FL experiments exhibit a reduction in recall by approximately 10\%. We posit that this substantial decline in performance stems from clusters tending to overfit the local distributions, thereby diminishing the meaningfulness of aggregation compared to training with all clients collectively. 
Concerning the server optimizers, AdaGrad demonstrates slightly superior performance compared to others. As a result, we opt to employ the standard SGD without momentum for the other baselines. 
    
\subsection{Data Quantity Skewness in FedVPR} 

\begin{table}
\caption{Addressing the clients' quantity heterogeneity. We compare the R@1 (\%) of the FedAvg baseline (grey background) with the ones of FedAvg and FedVC \citep{hsu2020federated} with a fixed number of iterations per client per round. $B$ is the local mini-batch size.}
\label{tab:fl-iterations}
\centering
\begin{adjustbox}{width=\linewidth}
\footnotesize
    \begin{tabular}{cccc}
        \toprule
        \textbf{Local Iterations} & \textbf{Rounds} & \textbf{FedAvg} & \textbf{FedVC} \\
        \midrule
        \rowcolor{tablegrey!20} $\min \left ( \left \lfloor \nicefrac{|\mathcal{D}_k|}{B} \right \rfloor, 2500 \right )$ & 300 & 61.0 \scriptsize{$\pm$ 0.6} & - \\
        125 & 3200 & 66.6 \scriptsize{$\pm$ 0.8} &  62.3 \scriptsize{$\pm$ 1.1}\\
        250 & 1600 & 66.0 \scriptsize{$\pm$ 1.7} &  65.9 \scriptsize{$\pm$ 1.0}\\
        500 & 800 & 66.4 \scriptsize{$\pm$ 1.6} &  \textbf{67.7 \scriptsize{$\pm$ 0.4}} \\
        1000 & 400 & 61.7 \scriptsize{$\pm$ 2.4} &  66.8 \scriptsize{$\pm$ 0.5}\\
        2000 & 200 & 58.8 \scriptsize{$\pm$ 1.8} &  65.2 \scriptsize{$\pm$ 0.9}\\
        4000 & 100 & 57.3 \scriptsize{$\pm$ 2.5} &  60.6 \scriptsize{$\pm$ 1.0}\\
        \bottomrule
    \end{tabular}

\end{adjustbox}
\vspace{-10pt}
\end{table}

\Cref{tab:splits} highlights significant variations in the number of sequences or images among clients (\ie, \textit{quantity heterogeneity}), particularly evident in the Proximity split with a radius of 2000m - our reference federated dataset. This section analyzes how this phenomenon affects the final performance.

In heterogeneous settings, an increased number of local training steps (updates within a client over a batch of data) fosters client drift and destructive interference during aggregation \cite{karimireddy2021scaffold,caldarola2022improving}. Thus, a larger local dataset leads to more updates and potentially negatively impacts the training process. Motivated by these insights, \cref{tab:fl-iterations} investigates how the data quantity skewness and the number of local training iterations affect performances of algorithms trained within the FedVPR framework, focusing on FedAvg and the state-of-the-art algorithm FedVC (Federated Virtual Clients) \cite{hsu2020federated}. FedVC specifically addresses variations in client data sizes by splitting large datasets into smaller clients and replicating smaller ones. This ensures all participating \textit{virtual} clients contribute roughly the same amount of data during each training round. To prevent knowledge loss, larger clients are resampled with higher probability. 

Given a fixed amount of total iterations $I_{tot}$, we either vary the local iterations $I_{loc}$, or the  training rounds $T$ such that $I_{loc}\times T\times|\mathcal{C}^t|=I_{tot}$. With larger $I_{loc}$, smaller datasets are used multiple times within a client, while fewer iterations might lead to an incomplete view of larger datasets. We set $I_{tot}=2,000,000$, $|\mathcal{C}|^t=5$ and vary $I_{loc}$ and $T$. 

This analysis reveals that reducing the influence of data imbalances can improve performance by up to 5\%. However, there exists a trade-off between communication rounds and final accuracy. As shown in \cref{tab:fl-iterations}, increasing $T$ (more communication) while reducing the local steps leads to performance improvement. Conversely, excessively increasing local computation at the expense of the number of rounds deteriorates the final performance. Finally, FedVC's sampling strategy, which favors larger clients, consistently improves performance when each device performs more than 500 local updates per round. However, for fewer local updates (125 and 250), FedVC shows a decrease in accuracy.

\subsection{Heterogeneity of Local Augmentations}
\label{sec:augm}

\begin{table}[]
\caption{\textbf{Augmentation}. Comparison of data augmentation strategies. The baseline represents training without augmentation, while the Client-specific color jitter strategy aims at simulating the system heterogeneity typical of FL \cref{sec:augm}.}
\centering
\footnotesize
    \begin{tabular}{llr}
        \toprule
            \textbf{Augmentation} & \textbf{R@1 (\%)} \\
        \midrule
        \rowcolor{tablegrey!20}
            Baseline & 61.0 \scriptsize{$\pm 0.6$} \\
        \midrule
            Client-specific color jitter & 53.5 \scriptsize{$\pm 2.5$} \\
            Color jitter & 64.7 \scriptsize{$\pm 0.9$} \\
            Color jitter + random resize crop & 65.7 \scriptsize{$\pm 0.6$} \\
        \bottomrule
    \end{tabular}
    \label{tab:fl-augmentation}
\end{table}

\Cref{tab:fl-augmentation} illustrates the performance variations with different levels of data augmentation during training using FedAvg. 
Varying levels of color jitter among clients (\textit{client-specific color jitter}) simulate significant statistical heterogeneity. This heterogeneity could arise from various sources, as client devices capture data with varying camera qualities and under diverse environmental conditions (lighting, weather). 
As expected, this experiment results in a severe performance degradation ($-7.5$ points w.r.t. the baseline). To isolate the effects of color jitter augmentation from statistical heterogeneity, we apply the same color jitter augmentation uniformly across all clients, revealing an improvement of nearly 4 percentage points in R@1 w.r.t. the baseline.
An additional random resized crop yields an increase in performance of 1 percentage point. We posit that the substantial benefits observed with stronger data augmentation primarily stem from the relatively small local datasets. Without robust augmentation, clients tend to overfit on local data, culminating in a meaninglessly aggregated model at the end of each round.

\subsection{Impact of Data Distribution on Local Mining}
As discussed in \cref{sec:centr_vpr}, the mining algorithm is a crucial factor affecting performances in VPR.
In centralized training, the model can access the entire database to select negative examples. On the other hand, in FL, clients can only rely on their local collection of images, which come from a limited geographical area, thus limiting the negative sampling distribution.
To study the extent to which this limitation represents an issue for FL, we run centralized experiments with limited available images during mining (\cref{tab:centralized-local-mining}). 
For each query, the server accesses negative images only within the closest $N$ database sequences.

\begin{table}
    \caption{\textbf{Local mining}. In a centralized scenario, we constrain the mining procedure to select negatives within the closest database sequences, rather than the global database, to study its effect. This setting emulates a federated scenario.}
    \vspace{-0.4cm}
    \begin{center}
    {{
    \begin{adjustbox}{width=.7\linewidth}
        \begin{tabular}{llllr}
            \toprule
                \textbf{Mining} & \textbf{\# Sequences} &  \textbf{DB size} & \textbf{R@1} \\
            \midrule
            \rowcolor{tablegrey!20}
                Baseline & - &  1k & 66.0 \scriptsize{$\pm 0.4$} \\
                \midrule
                Local & 333 &  28k & \textbf{68.8} \scriptsize{$\pm \textbf{0.2}$} \\
                Local & 20 & 3k & 58.0 \scriptsize{$\pm 0.9$} \\
            \bottomrule
        \end{tabular}
    \end{adjustbox}
    }}
    \end{center}
    \vspace{-0.4cm}
    \label{tab:centralized-local-mining}
\end{table}
The results confirm our expectation that focusing on an overly restricted geographical range of images can hinder performance in VPR. Interestingly, having an extremely wide geographical spread of images doesn't necessarily lead to better learning either.
Limiting the image range to a moderate level ($\approx333$ sequences, \ie, neighborhood to city scale) leads to a slight performance improvement. This suggests that focusing on a  localized area can be beneficial for VPR, motivating FedVPR. However, excessively restricting the range to a very local level (20 sequences) proves detrimental, resulting in performance even lower than a FL approach (cf. \cref{tab:fl-cmp}). This is likely because most clients in the federated setting have access to a wider variety of images.

These findings challenge the traditional assumption that geographical diversity is essential for VPR and suggest that a balance might exist between geographical scope and training data diversity for optimal VPR results.
\section{Conclusion}
In this work, we introduced \textbf{FedVPR}, a novel Federated Learning framework specifically designed for Visual Place Recognition (VPR) tasks. This approach addresses the growing need for distributed VPR solutions in applications like autonomous vehicles and mobile augmented reality. We analyzed the unique challenges of federated VPR compared to classification tasks and demonstrated that FedVPR can achieve performance comparable to a centralized model while minimizing resource consumption on individual devices. Our exploration identified key design choices impacting the performance-cost trade-off, including the number of local iterations, data augmentation strategies, and image resolution. These simple yet effective tools effectively mitigate statistical heterogeneity, validating the feasibility of the FedVPR setting. We believe this work opens a new avenue for VPR research while offering a realistic and valuable task for federated learning research.

\paragraph*{Acknowledgments.}
This project was supported by CINI (Consorzio Interuniversitario Nazionale per l'Informatica). Computational resources were provided by HPC@POLITO.

{
    \small
    \bibliographystyle{ieeenat_fullname}
    \bibliography{biblio}
}

\clearpage
\appendix
\section*{Appendix}

The Appendix is organized as follows:
\begin{itemize}
    \item \cref{app:centralized}: additional details on centralized runs.
    \item \cref{app:impl_details}: implementation details.
    \item \cref{app:plots}: additional analyses on MSLS Proximity split.
    \item \cref{app:hfl}: ablation studies on Hierarchical Federated Learning.
\end{itemize}

\section{Centralized runs}
\label{app:centralized}
This section introduces additional details on the centralized experiments presented in \cref{exp:centr}. 
Centralized training continues until the network performance plateaus for five consecutive epochs. This approach results in training different networks for a varying number of epochs depending on their convergence speed.
\cref{tab:centralized-time} compares the selected model architectures in terms of number of epochs, recall and training time. Based on these results and on the number of parameters (\cref{tab:centralized-baseline}), we select ResNet18 truncated as our network. 
\begin{table}[h]
    \caption{\textbf{Centralized baselines}: model architectures compared in terms of number of epochs, recall (R@1) and training time in the centralized scenario.
    }
    \centering
    \footnotesize
            \begin{tabular}{lccc}
                \toprule
                \textbf{Backbone}  & \textbf{Epochs} & \textbf{R@1} & \textbf{Time} \\
                \midrule
                ResNet18 truncated & 40 & 42.9 $\pm$ \scriptsize{2.5} &15h30 \\
                ResNet18 & 31 & 60.1 $\pm$ \scriptsize{0.3}& {10h15} \\
                VGG16 & 30 & 46.3 $\pm$ \scriptsize{0.5}&  28h30 \\
                \bottomrule
            \end{tabular}

    \label{tab:centralized-time}
\end{table}

\section{Implementation details}
\label{app:impl_details}
This section extends \cref{sec:impl_details} with additional implementation details on our experiments. The codebase  is written in Python with PyTorch for neural networks optimization. The experiments are run on the Nvidia Titan X GPU with 12GB of VRAM. All runs are averaged across 3 different seeds.

\myparagraph{Model} The experiments are run using a ResNet18 truncated after the third convolutional layer. The pooling layer is GeM except for the baseline experiments available in \cref{tab:centralized-baseline} where we tested SPOC and MAC as well. The image resolution is always 288x384 pixels except for the cases in \cref{tab:centralized-baseline} and  \cref{fig:centralization-resolution} where we tried different values: 96x128, 192x256, 384x512, and 480x640 pixels.

\myparagraph{FL baselines} The number of rounds $T$ in the FL experiments is set to 300, with 5 clients per round. The server  optimizer is always SGD with learning rate 1, \ie, FedAvg. \cref{tab:app_fed_lr} reports the hyperparameters used for the ablation on the server-side optimizers (\textsc{ServerOpt} in \cref{sec:fedvpr}) from \cref{tab:fl-cmp}. 
In local training, the optimizer is Adam with learning rate is always set to $1e-5$ and momentum 0. Each client runs one epoch. Unless otherwise specified, the maximum number of local iterations is set to 2500. When comparing H-FL with FedAvg (\cref{tab:fl-cmp}), we modify the number of rounds $T$ and participating clients $C$ accordingly, as summarized in \cref{tab:app_rounds}.

\begin{table}[]
    \centering
    \caption{Server-side optimizers hyperparameters (learning rate $\eta_s$ and momentum $\beta_s$).}
    \footnotesize
            \begin{tabular}{lcc}
                \toprule
                    \textbf{Method}    & \boldmath$\eta_s$   & \boldmath$\beta_s$ \\
\midrule
FedAvg & 1 & 0\\
{FedSGD}     & { 0.1} & { 0.9}                                   \\
{FedAdam}    & { 0.1} & { 0.9}                                   \\
{FedAdaGrad} & { 0.01} & { 0.9}                                   \\
                \bottomrule
            \end{tabular}
    \label{tab:app_fed_lr}
\end{table}

\begin{table}[]
    \centering
    \caption{Number of rounds $T$ and selected clients per round $C$ when comparing FedAvg with H-FL.}
    \footnotesize
            \begin{tabular}{lcc}
                \toprule
                        \textbf{Method} & \boldmath$T$   & \boldmath$C$ \\
                        \midrule
                        FedAvg     & 75   & 20 \\
                        H-FL Continent & 75    & 20 (4 continents by 5 clients per continent) \\
                        \midrule
                        FedAvg        & 15  & 105 \\
                        H-FL City      & 15     & 105 (21 cities by 5 clients per city) \\
                \bottomrule
            \end{tabular}
    \label{tab:app_rounds}  
\end{table}

\myparagraph{Data augmentation} We study the effect of data augmentation techniques in \cref{sec:augm}. Due to the required increased time, data augmentation is not used by default in the other experiments. In \cref{sec:augm}, data augmentation is  applied with 50\% probability. We apply color jitter (hue, saturation, brightness, and contrast) and random resize crop. 
Normalization instead is always applied with standard ImageNet values.

\myparagraph{Local mining} We set the number of sequences for computing the mining dataset to 333 and 20 and the number of images selected per sequence to 3 and 50 respectively.

\section{Distribution of clients in federated MSLS}
\label{app:plots}
\begin{figure*}[]
    \centering
    \subfloat[][Distribution of clients per \textbf{city}]{\includegraphics[width=.45\textwidth]{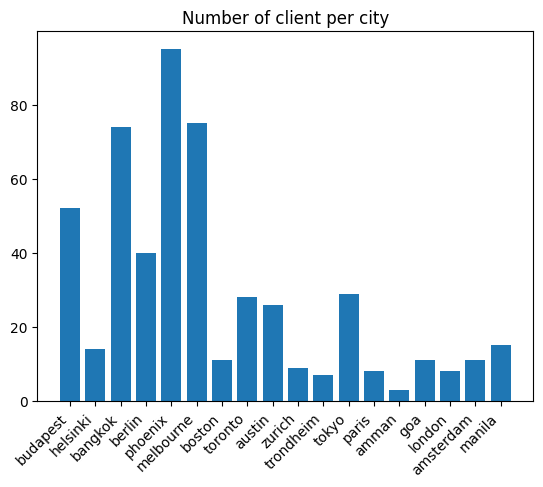}\label{fig:client-city}}
    \hfill
     \subfloat[][Distribution of clients per \textbf{continent}]{\includegraphics[width=.5\textwidth]{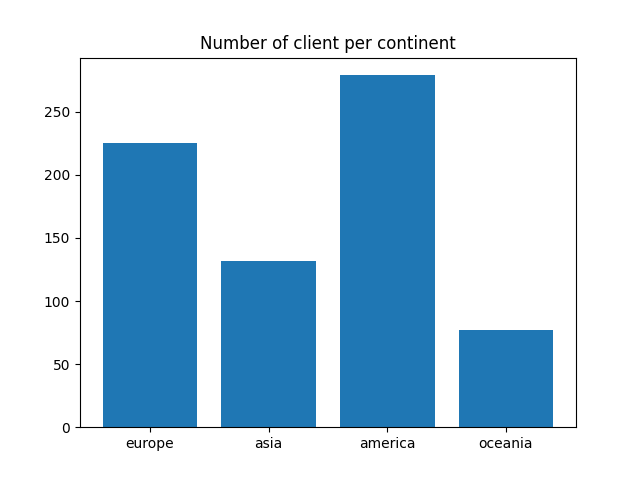}\label{fig:client-continent}}
    \caption{Clients distribution in the MSLS Proximity split.}  
\end{figure*}

\begin{figure*}[]
    \centering
    \subfloat[][Distribution of images per \textbf{city}]{\includegraphics[width=.45\textwidth]{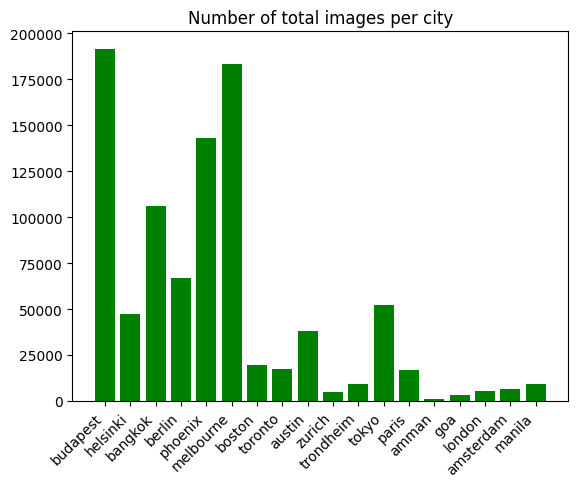}\label{fig:client-images}}
    \hfill
     \subfloat[][Distribution of images per \textbf{continent}]{\includegraphics[width=.5\textwidth]{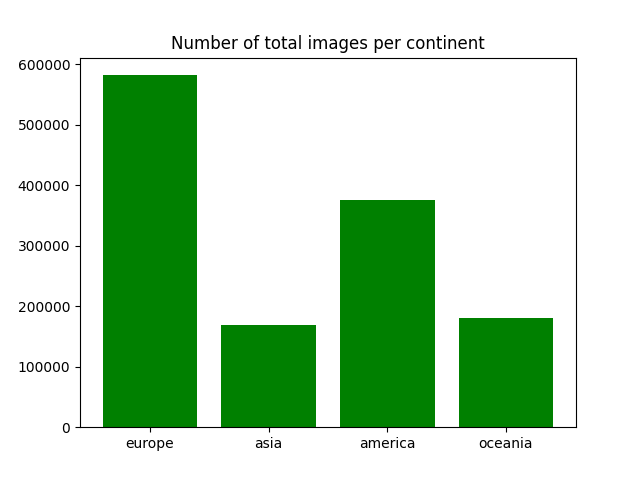}\label{fig:continent-images}}
    \caption{Images distribution in the MSLS Proximity split.}  
\end{figure*}

In this section, we present additional analyses on the MSLS federated splits described in \cref{sec:dataset}. 
Focusing on the \textit{Proximity} split, \cref{fig:client-city} shows the distribution of clients across cities. We note that Budapest, Bangkok, Phoenix and Melbourne are the most populated. \cref{fig:client-images} shows that those same cities are also the ones containing most images.  \cref{fig:client-continent,fig:continent-images} repeat the same analyses per continent: even if most of the clients are found in America, Europe has most of the images, while the least populated continent is Oceania but Asian clients have in total less images.

\section{Ablation studies on H-FL}
\label{app:hfl}
\cref{tab:fl-aggregation} investigates the effect of the round interval ($T_s$) between aggregation steps in H-FL for continent-based clustering. The results show that an optimal value exists for $T_s$.  Setting $T_s$ too low hinders the cluster-specific models from learning generalizable information, while a very high $T_s$ leads to reliance on outdated updates.  The best performance is achieved with $T_s=15$.

\begin{table}
\caption{\textbf{Comparison of different aggregation interval $T_s$ in h-FL.} The experiments are run with the continental aggregation with 5 clients per continent at each round and carried on for 300 rounds.}
\label{tab:fl-aggregation}
\centering
\footnotesize
    \begin{tabular}{cc}
        \toprule
        \boldmath$T_s$ & \textbf{R@1} \\
        \midrule
        5 & 60.1  \scriptsize{$\pm$ 0.8}\\
        10 & 60.4  \scriptsize{$\pm$ 0.9}\\
        15 & \textbf{61.1} \scriptsize{$\pm$ \textbf{0.6}}\\
        20 & 60.0  \scriptsize{$\pm$ 0.9}\\
        25 & 60.3  \scriptsize{$\pm$ 0.4}\\
        \bottomrule
    \end{tabular}
\end{table}

\end{document}